\pdfoutput=1 

\documentclass[lettersize,journal]{IEEEtran}
\usepackage{amsmath,amsfonts}
\usepackage{algorithmic}
\usepackage{algorithm}
\usepackage{array}
\usepackage[]{subfig}
\usepackage{textcomp}
\usepackage{stfloats}
\usepackage{url}
\usepackage{verbatim}
\usepackage{graphicx}
\usepackage{cite}
\usepackage{hyperref}
\usepackage{bm}
\usepackage{makecell}
\usepackage{footnote}
\usepackage{float}
\usepackage{subfig}
\usepackage{ragged2e}
\usepackage{amsthm}
\usepackage{dsfont}
\usepackage{lipsum} 
\usepackage{booktabs}
\usepackage{multirow}
\usepackage{xcolor}
\usepackage{makecell}

\usepackage{enumitem} 
\usepackage{pifont}

\makeatletter

\newcommand{\Rmnum}[1]{\expandafter\@slowromancap\romannumeral #1@}
\makeatother

\def\x{\boldsymbol{x}}

\theoremstyle{definition}

\usepackage{cuted}

\begin{document}

\title{Attention-embedded Quadratic Network (Qttention) for Effective and Interpretable Bearing Fault Diagnosis}

\author{Jing-Xiao~Liao$^{1}$, Hang-Cheng Dong$^{1}$, Zhi-Qi Sun$^{1}$,  Jinwei Sun$^{1}$,  Shiping Zhang$^{1*}$, Feng-Lei Fan$^{2*}$
\thanks{*Shiping Zhang (spzhang@hit.edu.cn) and Feng-Lei Fan (hitfanfenglei@gmail.com) are co-corresponding authors.}
\thanks{$^{1}$Jing-Xiao Liao,  Hang-Cheng Dong, Zhi-Qi Sun, Jinwei Sun, Shiping Zhang are with School of Instrumentation Science and Engineering, Harbin Institute of Technology, Harbin, China.}
\thanks{$^{2}$Feng-Lei Fan is with Weill Cornell Medicine, Cornell University, New York City, USA.}}



\maketitle

\begin{abstract}
Bearing fault diagnosis is of great importance to decrease the damage risk of rotating machines and further improve economic profits. Recently, machine learning, represented by deep learning, has made great progress in bearing fault diagnosis. However, applying deep learning to such a task still faces a major problem. A deep network is notoriously a black box. It is difficult to know how a model classifies faulty signals from the normal and the physics principle behind the classification. To solve the interpretability issue, first, we prototype a convolutional network with recently-invented quadratic neurons. This quadratic neuron empowered network can qualify the noisy bearing data due to the strong feature representation ability of quadratic neurons. Moreover, we independently derive the attention mechanism from a quadratic neuron, referred to as qttention, by factorizing the learned quadratic function in analogue to the attention, making the model with quadratic neurons inherently interpretable. Experiments on the public and our datasets demonstrate that the proposed network can facilitate effective and interpretable bearing fault diagnosis.
\end{abstract}


\section{Introduction}
\IEEEPARstart{T}{he} reliability of rotating machines such as wind turbines and aircraft engines is an critical issue and must be ensured at all times in the industrial field. Among major mechanical components of a rotating machine, bearings are the most popular source of faults. According to several studies \cite{bonnett2008increased}, bearing faults are responsible for 40 to 70 percents of electromagnetic drive system failures. Therefore, the diagnosis to bearing faults is of great criticality to improve the availability of rotating machines and further avoid economic loss. A common and viable method to detect bearing faults is analyzing vibration signals that are produced by attaching the measuring instrument to the rotating bearing \cite{mcfadden1984model}. Previous diagnosis works can be divided into two categories: signal processing-based methods vs data-driven methods \cite{ding2022self}. 

Signal processing-based methods usually utilize Fourier transform \cite{pandarakone2018deep}, short-time Fourier transform \cite{gao2015feature}, Hilbert-Huang transform \cite{rai2007bearing,osman2016morphological}, etc. to transform a sequence of a signal into the frequency domain or time-frequency domain. Then, the spectral analysis is conducted to capture the feature of failed faults from the normal. Despite theoretical soundness, these methods suffer from the following weaknesses: 1) Because bearings are always installed in a complex mechanical system, the measured signal is filled with interference from other mechanical components and background noise from the measurement device \cite{bonnett2008increased}. This requires a sophisticated and careful design of the denoising model in order to extract various fault characteristics from the complex noise environment. 2) Because different bearing structures and diameters lead to different characteristic frequencies of the faults, the manually-determined parameters are required in performing signal processing-based methods to maximize their diagnosis performance, which is neither intelligent nor scalable due to the complexity of industrial fields \cite{ding2022self}. 

To overcome the above weaknesses, machine learning, particularly deep learning, is widely introduced for bearing fault diagnosis with the promise of big data empowered end-to-end accurate feature discrimination. The past several years have seen a variety of deep learning models developed. Wen \textit{et al.} \cite{wen2017new} first folded 1D time-domain signals into 2D images, and then used LeNet-5 as a backbone to classify these images. Concurrently, more studies directly worked on 1D time-domain signals with the help of 1D-CNN \cite{wen2017new, zhao2019deep}. WDCNN adopted a wide convolution kernel in the first layer as the signal extractor for 1D signals and achieved satisfactory results on several bearing fault benchmarks \cite{zhang2017new}. Due to the outperformance of the WDCNN, more advanced models have been devised based on the WDCNN \cite{wang2019understanding, shenfield2020novel}. Moreover, the attention module was introduced in bearing fault diagnosis because it helps a network focus on fault features and further enhances feature extraction capabilities. Researchers in \cite{li2019understanding,wang2019understanding, yang2020interpreting, wang2020intelligent} inserted a multi-attention module into a 1D-CNN, which not only improves the network's classification accuracy but also makes the network interpretable.

Despite that deep learning models have made a big stride in bearing fault detection, there are still challenges ahead that need to be addressed. 1) (Effectiveness) Currently, deep learning models can perform superbly, \textit{i.e.}, achieving over 90\% accuracy, when the signals are clean or big data are curated. However, when signals are highly corrupted by noise, the performance of deep learning models drops dramatically.
The above issue should be accommodated if one wants to implement a deep learning model in industrial fields. 2) (Interpretability) Deep learning, comprising of a series of nonlinear or recursive operations, is notoriously a black box \cite{zhang2018visual}. Due to the lack of interpretability, it is hard to know the physics principle used by a model to pick fault signals out and what changes can be made to further boost the model's diagnosis performance. 

\begin{figure*}[t]
    \centering
    \includegraphics[width=0.85\linewidth]{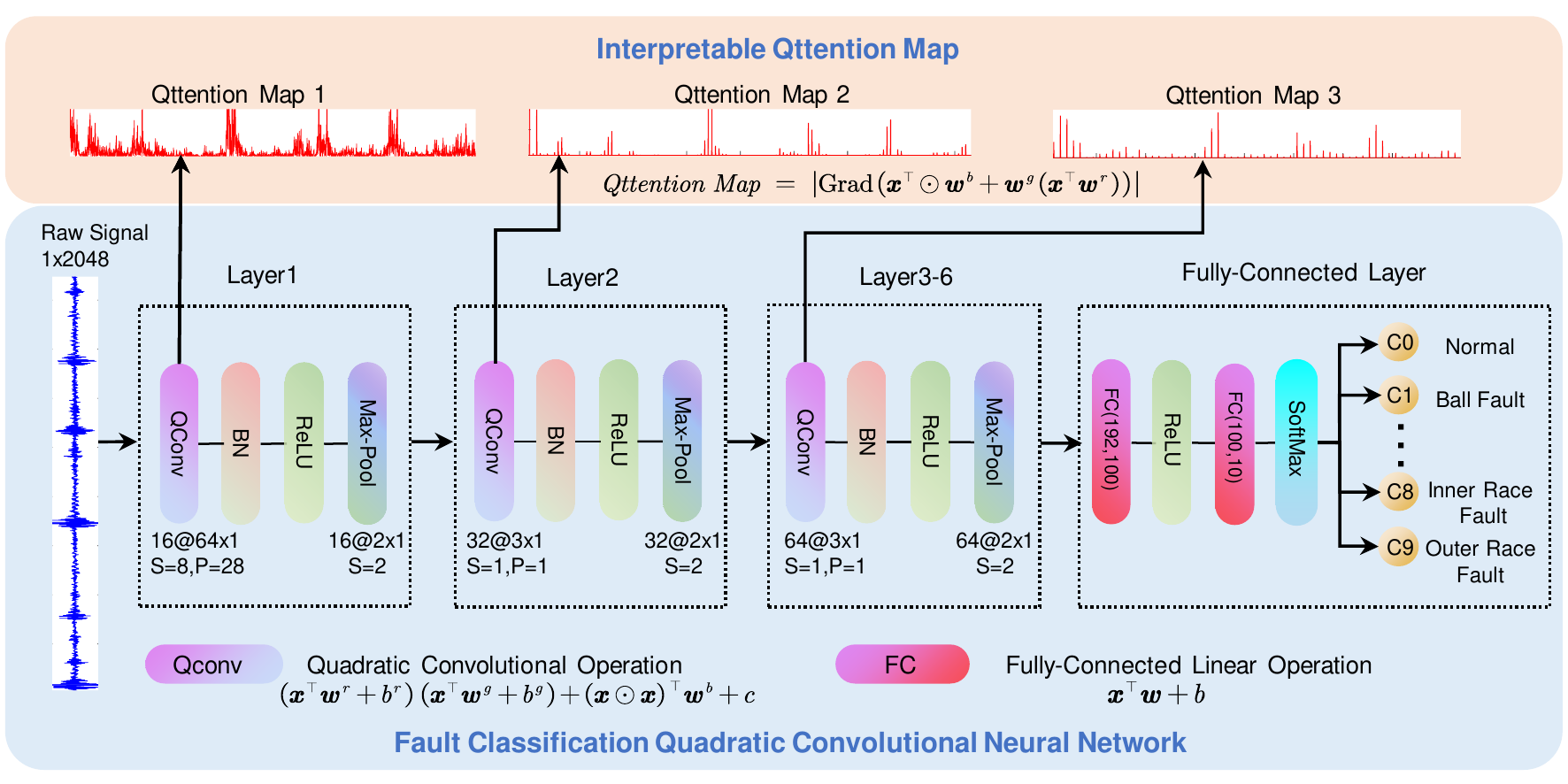}
    \caption{The proposed quadratic convolutional neural network (QCNN).}
    \label{fig:qcnn}
    \vspace{-0.5cm}
\end{figure*}

To deal with the challenges in effectiveness and interpretability, here we turn our eyes to the quadratic neurons and associated models. Recently, motivated by introducing neuronal diversity in deep learning, a new type of neuron called quadratic neuron \cite{b4} was proposed by replacing the inner product in the conventional neuron with a simplified quadratic function. \textit{Why are quadratic neurons promising in solving the effectiveness and interpretability issues?} For the former, it has been empirically and theoretically shown \cite{b4, fan2021expressivity} that a quadratic network enjoys a high feature extraction ability. Due to such a character, quadratic models have the potential of qualifying the noisy bearing diagnosis. For the latter, as our independent methodological finding, the learned quadratic function in a quadratic neuron can induce an attention map through factorization, referred to as \textit{qttention}, making a quadratic network interpretable. Compared to the classical attention mechanism, which is global and computationally expensive, the qttention spotlights local fine-grained importance maps and consumes fewer parameters, which is a valuable addition to the attention-based interpretability. In summary, our contributions are twofold: 

\begin{itemize}
 
 \item We propose a simple and effective model that is made of quadratic neurons for bearing fault diagnosis. The quadratic neurons directly augment the model to outperform other state-of-the-arts in noisy data settings. Different from previous structural modifications, our innovation is at the neuronal level, applying different neurons to bearing fault diagnosis and upgrading CNNs-based bearing fault diagnosis models.

\item Methodologically, we derive the qttention mechanism from quadratic neurons, which greatly facilitates the understanding to the model made of quadratic neurons, with an emphasis on the model's physics mechanism when combined with envelope spectrum analysis.

\end{itemize}

\section{Methodology}
\subsection{Quadratic Convolutional Neural Network}

\underline{Quadratic neuron:} Encouraged by the neuronal diversity in a biological neural system, a new type of neuron called quadratic neuron \cite{b4} was proposed to promote the neuronal diversity in deep learning. A quadratic neuron \cite{b4} integrates two inner products and one power term of the input vector before nonlinear activation. Mathematically, suppose that the input vector is $\boldsymbol{x} = (x_1, x_2, \ldots ,x_n)^\top$, the output $f(\boldsymbol{x}): \mathbb{R}^n \to \mathbb{R}$ of a quadratic neuron is 
\begin{equation}
\scalebox{0.9}{
$
\begin{aligned}
\sigma(f(\boldsymbol{x})) = &\sigma\Big((\sum_{i=1}^{n} w_{i}^r x_i +b^r)(\sum_{i=1}^{n} w_{i}^g x_i +b^g) + \sum_{i=1}^{n} w_{i}^b x_{i}^2+c \Big) \\
=&\sigma\Big((\boldsymbol{x}^\top\boldsymbol{w}^{r}+b^{r})(\boldsymbol{x}^\top\boldsymbol{w}^{g}+b^{g})+(\boldsymbol{x}\odot\boldsymbol{x})^\top\boldsymbol{w}^{b}+c\Big),
\end{aligned}$}
\label{qneq1}
\end{equation}
where $\sigma(\cdot)$ is a nonlinear activation function, $\odot$ denotes the Hadamard product, $\boldsymbol{w}^r,\boldsymbol{w}^g, \boldsymbol{w}^b\in\mathbb{R}^n$ are weight vectors, and $b^r, b^g, c\in\mathbb{R}$ are biases. Superscripts $r,g,b$ are just marks for convenience without special implications. 

With the neuronal diversity, we argue that the network design consists of two parts: the neuronal design and the structural design. 
Figure \ref{fig:qcnn} shows the proposed quadratic convolutional neural network (QCNN). At the neuronal level, we adopt quadratic neurons in hope that they can facilitate bearing diagnosis via the strong feature representation power. Specifically, the QCNN substitutes the conventional convolution with the quadratic convolution operations in convolutional layers. Structurally, we inherit the structure of the WDCNN \cite{zhang2017new} as the structure of the proposed model because the WDCNN is a well-established model whose design has been considered by many follow-up studies. This structure stacks 6 CNN blocks and 1 fully-connected layer.

\underline{Superiority of representation:} The quadratic neurons have the intrinsic enhancement in terms of representation ability, \textit{i.e.}, the enhancement is not due to the increased parameters but the involved nonlinear computation. First, the nonlinear mapping of a conventional neuron is only provided by activation function, whereas a quadratic neuron adds an extra non-linear mapping by using the quadratic aggregation function. As such, a single quadratic neuron can realize XOR logic which is not doable by the conventional neurons unless they are connected into a network.
Second, a quadratic neuron is not equivalent to the combination or summation of three conventional neurons. Suppose that the activation function $\sigma(\cdot)$ is ReLU, the combination of conventional neurons can only be a piecewise linear function, but a quadratic neuron is a piecewise polynomial function \cite{fan2021expressivity}. As we know, a polynomial spline is better at approximating complicated functions than a linear spline.


\underline{Training strategy:} With the quadratic network as a more powerful model, a problem naturally arises: how can we ensure that the model can find a better solution? To address this problem, Fan \textit{et al.} \cite{fan2021expressivity} proposed the so-called ReLinear algorithm, where the parameters in a quadratic neuron are initialized as $\boldsymbol{w}^g=0, \boldsymbol{b}^g=1, \boldsymbol{w}^b=0,\boldsymbol{c}=0$, and $\boldsymbol{w}^r, b^r$ follow the normal initialization. Consequently, during the initialization stage, every quadratic neuron in a network degenerates to a conventional neuron. Next, during the training stage, a normal learning rate $\gamma_r$ is cast for $(\boldsymbol{w}^r, b^r)$, and a relatively small learning rate $\gamma_{g,b}$ for $(\boldsymbol{w}^g, b^g,  \boldsymbol{w}^b,c)$. 


\vspace{-0.2cm}
\subsection{Efficient Neuron-induced Attention}

The attention module \cite{chaudhari2021attentive} was originally derived for sequence tasks such as natural language processing. 
Mathematically, given the input signal $\x \in \mathbb{R}^{n\times k}$, the commonly-used scaled dot-product self-attention \cite{luong2015effective} is formulated as 
\begin{equation}
    \mathrm{softmax}\left(\x \boldsymbol{W^Q} (\x \boldsymbol{W^K})^\top/\sqrt{d_k}\right) (\x \boldsymbol{W^V}),
    \label{eq:alignfun}
\end{equation}
where $\boldsymbol{W^Q}, \boldsymbol{W^K}, \boldsymbol{W^V} \in \mathbf{R}^{k\times k'}$ are the projections for the query, key, and value, respectively, and $d_k$ is the scaling factor.
The attention that reflects the importance of the input is taken as the following formula from \eqref{eq:alignfun}:
\begin{equation}
Att(\x)=\mathrm{softmax}\left(\x \boldsymbol{W^Q} (\x \boldsymbol{W^K})^\top/\sqrt{d_k}\right) .
\end{equation}

The above attention mechanism is widely adopted in the transformer models \cite{liu2021swin}. Concurrently, channel and spatial attention modules are developed to adapt the CNN \cite{woo2018cbam}:
\begin{equation}
\begin{aligned}
\boldsymbol{F}^{\prime} &=\boldsymbol{M}_{\boldsymbol{c}}(\boldsymbol{F}) \odot \boldsymbol{F} \\
\boldsymbol{F}^{\prime \prime} &=\boldsymbol{M}_{\boldsymbol{s}}\left(\boldsymbol{F}^{\prime}\right) \odot \boldsymbol{F}^{\prime},
\end{aligned}
\label{eq:convatt}
\end{equation}
where $\boldsymbol{F} \in \mathbb{R}^{C\times H \times W}$ denotes an intermediate feature map produced by a convolutional layer, $\boldsymbol{M}_{\boldsymbol{c}} \in \mathbb{R}^{C\times 1 \times 1}$ denotes a 1D channel attention map, and 
$\boldsymbol{M}_{\boldsymbol{s}} \in \mathbb{R}^{1\times H \times W}$ denotes a 2D spatial attention map. Furthermore,
\begin{equation}
\begin{aligned}
\boldsymbol{M}_{\boldsymbol{c}}(\boldsymbol{F}) &=\sigma(\operatorname{MLP}(\operatorname{AvgPool}(\boldsymbol{F}))+\operatorname{MLP}(\operatorname{MaxPool}(\boldsymbol{F}))) \\
&=\sigma\left(\boldsymbol{W}_{\boldsymbol{1}}\left(\boldsymbol{W}_{\boldsymbol{0}}\left(\boldsymbol{F}_{\operatorname{{avg}}}^{\boldsymbol{c}}\right)\right)+\boldsymbol{W}_{\boldsymbol{1}}\left(\boldsymbol{W}_{\boldsymbol{0}}\left(\boldsymbol{F}_{{\operatorname{max}}}^{\boldsymbol{c}}\right)\right)\right)\\
\boldsymbol{M}_{\boldsymbol{s}}(\boldsymbol{F}') &=\sigma\left(\mathrm{Conv}([\operatorname{AvgPool}(\boldsymbol{F}') ; \operatorname{MaxPool}(\boldsymbol{F}')])\right),
\end{aligned}
\label{eq:sub-att}
\end{equation}
where $\sigma(\cdot)$ is the sigmoid function, $\operatorname{AvgPool}$ is the average pooling, $\operatorname{MaxPool}$ is the max pooling, $\operatorname{MLP}$ is the MLP layer, and $\mathrm{Conv}$ is the convolution filter. Regardless of the variants, the key of the attention mechanism is to induce an importance map regarding the input that forces the network to discriminatively take advantage of the input information.

Based on such an observation, we find that a quadratic neuron also contains an attention mechanism, referred to as \textit{qttention}, by factorizing the learned quadratic function in analogue to attention. Mathematically, we factorize
the expression of a quadratic neuron (Eq. \eqref{qneq1}) as follows (we remove constant terms for conciseness):
\begin{equation}
\begin{aligned}
&\sigma((\boldsymbol{x}^\top\boldsymbol{w}^{r}+b^{r})(\boldsymbol{x}^\top\boldsymbol{w}^{g}+b^{g})+(\boldsymbol{x}\odot\boldsymbol{x})^\top\boldsymbol{w}^{b}+c) \\
=&\sigma (\boldsymbol{x}^{\top}\boldsymbol{w}^g\left( \boldsymbol{x}^{\top}\boldsymbol{w}^r+b^r \right) +b^g\boldsymbol{x}^{\top}\boldsymbol{w}^r+b^gb^r+\left( \boldsymbol{x}\odot \boldsymbol{x} \right) ^{\top}\boldsymbol{w}^b)\\
=&\sigma ( \boldsymbol{x}^{\top}\left( \boldsymbol{w}^g\left( \boldsymbol{x}^{\top}\boldsymbol{w}^r+b^r \right) \right) +\boldsymbol{x}^{\top}\left( \boldsymbol{w}^rb^g \right) 
+\boldsymbol{x}^{\top}\left( \boldsymbol{x}\odot \boldsymbol{w}^b \right) )\\
=&\sigma \left( \boldsymbol{x}^{\top}\left( \boldsymbol{x}\odot \boldsymbol{w}^b+\boldsymbol{w}^g\left( \boldsymbol{x}^{\top}\boldsymbol{w}^r+b^r \right) +\boldsymbol{w}^rb^g \right)  \right) \\
= & \sigma ( \boldsymbol{x}^{\top}( \underbrace{\boldsymbol{x}\odot \boldsymbol{w}^b+\boldsymbol{w}^g \boldsymbol{x}^{\top}\boldsymbol{w}^r}_{qttention} +\underbrace{\boldsymbol{w}^g b^r+\boldsymbol{w}^rb^g}_{bias} ) ) ,
\end{aligned}
\label{eq:quadratic}
\end{equation}
where we let
 \begin{equation}
     RawQtt(\boldsymbol{x})=\boldsymbol{x}^{\top}\odot \boldsymbol{w}^b+\boldsymbol{w}^g(\boldsymbol{x}^{\top}\boldsymbol{w}^r).
\label{eq:qtt}
 \end{equation}
 
We exclude the bias terms $\boldsymbol{w}^gb^r, \boldsymbol{w}^rb^g$ because they keep intact for different $\x$; therefore, they fall short of serving as importance scores. Furthermore, we calculate the gradient of $RawQtt(\boldsymbol{x})$ and take the absolute value to get the final qttention map:
 \begin{equation}
     Qtt(\x) = |\mathrm{Grad}(RawQtt(\boldsymbol{x}))|.
\label{eq:map}
 \end{equation}
 The reason of doing so is that the gradient can better indicate the trend of the change and further eliminate the common bias, which highlights significant areas of attention weights.
 
 \begin{figure}[htb]
    \centering
    \includegraphics{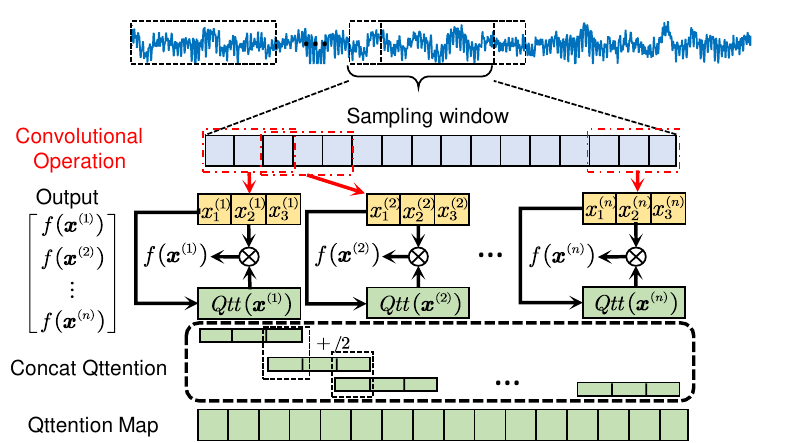}
    \caption{The workflow of establishing the qttention map for the entire signal in the framework of the convolutional operation.}
    \label{fig:qttention}
\end{figure}

In a convolutional layer, the qttention is tightly coupled to the convolution operation. Figure \ref{fig:qttention} illustrates the workflow of establishing the qttention map for the entire signal. Each convolutional kernel is considered to be a single neuron and a convolutional operation is shifting the kernel over the input with a constant stride. At each receptive field, the computation of the convolution kernel follows \eqref{qneq1} and generates a qttention map. The final qttention map for the entire signal is derived by concatenating all qttention maps at local receptive fields. When the stride is smaller than the length of the receptive field, two neighboring receptive fields overlap. Then, we average the qttention scores at the overlapped locations.

\textbf{Remark.} What makes the qttention different from the conventional attention module is that the conventional one usually is an independent plug-and-play module, but the qttention is neuron-induced. We summarize that the qttention has the following characteristics: 

\underline{Efficient:} The channel and spatial attention compute an element-wise product for the entire input.  Moreover, the channel attention module also includes an MLP network. Both the element-wise product for the entire input and the MLP network are computationally heavy. Assume a channel attention module (Eq. (\ref{eq:sub-att})) stacked after a convolutional layer $\boldsymbol{F} \in \mathbb{R}^{C \times H \times W}$, its number of parameters $P$ is
\begin{equation}
\begin{aligned}
P(\boldsymbol{M}_c(\boldsymbol{F}))&= 2P(\boldsymbol{W}_{\boldsymbol{1}_{(C \times \frac{C}{r})}}(\boldsymbol{W}_{\boldsymbol{0}_{\frac{C}{r} \times C}}(\boldsymbol{F^c}_{(C \times 1 \times 1)})))\\
&=2\times \frac{C^5}{r^2},
\end{aligned}
\end{equation}
where $C$ is an integer denotes the number of convolutional channels, and $r>0$ is a scale factor. For an entire block that contains a convolutional layer with an attention module, the number of parameters is at least 
\begin{equation}
    P(Att)=(C\times H\times W)+2\times \frac{C^5}{r^2}.
\end{equation}
Moreover, with the addition of the spatial attention module, the number of parameters will be larger.

In contrast, the qttention is generated by a convolution operation whose number of weights only scales with the kernel size, and parameters of qttention are the number of parameters of the quadratic convolutional layer:
\begin{equation}
    P(Qtt)=3 \times (C\times H \times W).
\end{equation}
The number of parameters of a qttention is far fewer than that of a typical attention module. Therefore, even if we use qttention in every convolutional layer, the efficiency is much higher than adding the attention module.

\underline{Local:} Compared to the typical attention module which is global, the qttention is local. The locality of the qttention is more suitable for bearing fault diagnosis. Usually, the attention module is applied after the signal is divided into several patches. Then, the attention score is assigned to each patch, and the attention map is for the entire signal but course-grained. In contrast, the qttention is associated with the convolution, which emerges at each local receptive field. The locality of the qttention fits the diagnosis because the fault usually presents in the time domain as periodic short-range high-amplitude vibrations.

\vspace{-0.2cm}
\section{Experiments}
In this section, we use two bearing fault datasets to investigate our proposed model. We first compare the QCNN with other SOTA methods under noisy conditions. The results show that the QCNN outperforms its competitors. Then, aided by the interpretability of qttention maps, we successfully decode the feature extraction process of the proposed model and the physics principle accounting for why the model can achieve good classification performance.

\subsection{Datasets Description}


\underline{CWRU bearing dataset}: This widely-used dataset \cite{9078761} is collected by Case Western Reserve University Bearing Data Center. Two deep groove ball bearings, 6205-2RS JEM SKF and 6203-2RS JEM SKF, are installed in the fan-end (FE) and drive-end (DE) of the electric motor. Single point defects with a diameter of 7, 14 and 21 mils are injected to outer race, inner race and ball of two bearings using electro-discharge machining. Therefore, this dataset has ten categories: nine types of faulty bearings and one healthy bearing. Specially, four levels (0HP, 1HP, 2HP, 3HP) of load are applied onto the shaft which slightly affects the motor speed (1797r/min, 1772/min, 1750r/min, 1730r/min). Vibration data are collected at two sampling rates: 12kHz and 48kHz for DE bearing faults. In this paper, we use the vibration signal collected at DE side with the 12kHz sampling rate. 
\begin{figure}[htbp]
\vspace{-0.5cm}
    \centering
    \includegraphics[width=\linewidth]{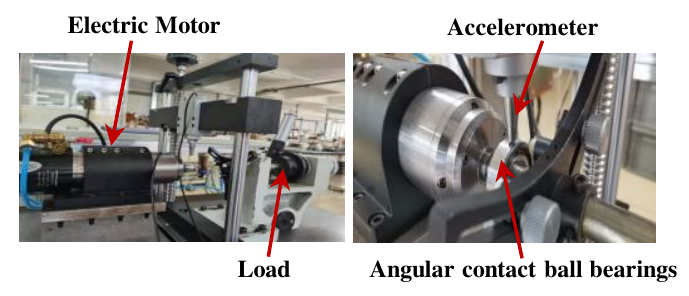}
    \caption{The rig for angular contact ball bearing test.}
    \label{fig:HITdataset}
\vspace{-0.5cm}
\end{figure}


\begin{table}[htbp]
\caption{Ten classes in our dataset. OR and IR denote that the faults appear at outer race and inner race, respectively.}
\centering
\begin{tabular}{|l|l|l|l|}
\hline
Label & Fault Mode        & Label & Fault Mode  \\ \hline
C1    & Health                                    & C6    & OR (Moderate)  \\ \hline
C2    & Ball cracking (Minor)       & C7    & OR (Severe)   \\ \hline
C3    & Ball cracking (Moderate)     & C8    & IR (Minor)   \\ \hline
C4    & Ball cracking (Severe)       & C9    & IR (Moderate) \\ \hline
C5    & OR cracking (Minor) & C10   & IR (Severe)   \\ \hline
\end{tabular}
\label{tab:class}
\end{table}

\begin{figure}
    \centering
    \includegraphics[width=\linewidth]{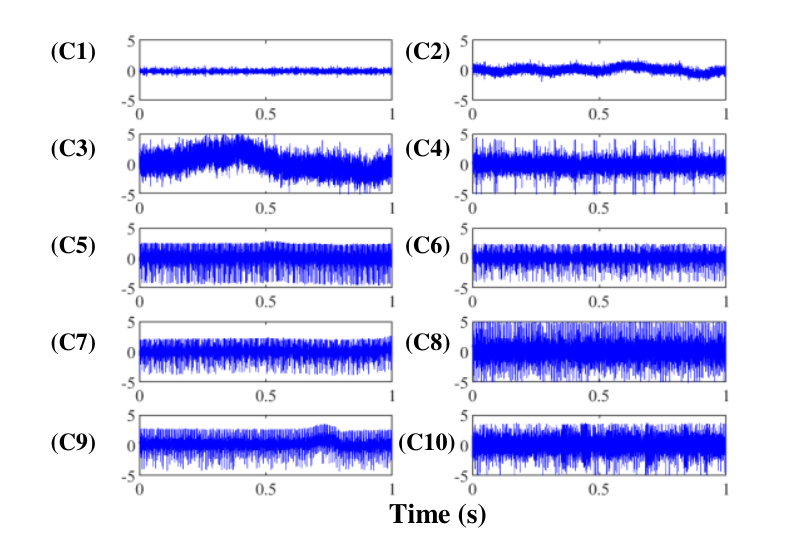}
    \caption{Raw signals with respect to ten classes of our dataset.}
    \label{fig:raw}
    \vspace{-0.7cm}
\end{figure}

\underline{HIT angular contact ball bearing crack dataset (HIT)}:
We conduct a bearing fault test in MIIT Key Laboratory of Aerospace Bearing Technology and Equipment, Harbin Institute of Technology. Figure \ref{fig:HITdataset} shows our bearing test rig. We utilize angular contact ball bearings HC7003 for the test. This bearing is for high speed rotating machines compared to a deep groove ball bearing. The accelerometer is directly attached to a bearing to collect vibration signals produced by bearings. Consistent with the CWRU dataset, we inject faults at the outer race (OR), inner race (IR), and ball with three levels (minor, moderate, severe). Table \ref{tab:class} summaries ten classes of bearings in our dataset. In the test, we choose the constant motor speed (1800 r/min) and use NI USB-6002 to acquire vibration signals with the 12kHz sampling rate. We record 47s of bearing vibration (561,152 points per category). Different from the CWRU dataset, bearing faults here are cracks of the same size but different depths, therefore, the vibration signals between different faults are more similar, making a diagnosis model more difficult to accurately classify. Figure \ref{fig:raw} shows the raw signals in the time domain with respect to ten categories of our dataset.

\begin{table}[b]
  \centering
  \caption{The summary of compared models properties. Time is the elapsed time to infer 1000 samples for each model.}
    \begin{tabular}{|l|c|c|c|}
    \hline
    Model    & \#Params & \#FLOPS & Inference Time \\
    \hline
    DCA-BiGRU & 626KB & 42.4M & 4.5977s \\
    \hline
    AResNet & 10.67MB & 260.4M & 14.7547s \\
    \hline
    RNN-WDCNN & 4407KB & 54.3M & 5.1949s \\
    \hline
    MA1DCNN & 4172KB & 299.7M & 5.5341s \\
    \hline
    WDCNN  & 284KB & 1.6M  & 0.1925s \\
    \hline
    QCNN  & 642KB & 4.7M  & 0.3581s \\
    \hline
    \end{tabular}%
  \label{tab:times}%
\end{table}%

\subsection{Experiment Setup}

\underline{Data preprocessing.} For our dataset, because the original signal is long, we randomly extract short sequences of 2048 numbers from each long sequence 1000 times. As such, we generate a total of 10000 samples. Then, all samples are normalized into $[-1, 1]$. Both datasets are split into training, validation, and test sets with a ratio of 0.5: 0.25: 0.25.  

\underline{Training settings.} The loss function is set to the cross-entropy loss, and we choose the stochastic gradient descent (SGD) \cite{bottou2012stochastic} optimizer to optimize both networks. To generically compare the performance of all models, we don't add extra training tricks, \textit{e}.\textit{g}., weight decay, learning rate decay. We train our model and baseline methods using grid search with the same hyperparameter search space. After searching, we set batch size = 64 for all methods. For the QCNN, the linear terms and quadratic terms are learned with different learning rates. Let $\gamma_{g,b}=\alpha\cdot \gamma_r$. We search $\gamma_r$ from $\{0.1, 0.3, 0.5, 0.8, 0.08, 0.05\}$ which is the same as all conventional methods and $\alpha$ from $\{10^{-1}, 10^{-2}, 10^{-3}, 10^{-4}, 10^{-5}\}$. 
 
We use the accuracy score to validate the performance of the proposed method, which is defined as
\begin{equation}
\nonumber
    {\rm Accuracy} = \frac{TP+TN}{TP + FN + FP + TN},
\end{equation}
where TP, TN, FP, and FN denote the number of true positive, true negative, false positive, and false negative, respectively. Furthermore, we add Gaussian noise into the raw input signal to verify model's performance under noise environments. The signal-to-noise ratio (SNR) is
$10 \log_{10}(P_s/P_n)$, 
where $P_s$ and $P_n$ are average power of the signal and noise.

All experiments are conducted in Windows 10 with an Intel i9 10900k CPU at 3.70 GHz and one NVIDIA RTX 3080Ti 12GB GPU. Our code is written in Python 3.8 with PyTorch, an open-source deep learning framework. We will share our code after this manuscript is published. 

\subsection{Classification Performance} 
The bearing signal is interfered with by other mechanical components and background noise from the measurement device, therefore, the noise-resistant capability of deep learning models is critical to the bearing fault diagnosis. We conduct experiments to verify the classification performance of the proposed model under noisy settings.

We compare our method with other SOTA methods, DCA-BiGRU \cite{zhang2022fault}, AResNet \cite{zhongdevelopment}, RNN-WDCNN \cite{shenfield2020novel}, MA1DCNN \cite{wang2019understanding}, WDCNN \cite{zhang2017new}. Because codes of all these counterpart models are publicly available, we replicate them by their code and utilize the same data preprocessing. Table \ref{tab:times} summarizes the sizes, \#FLOPS, and inference time of all models. Compared to competitors, the QCNN has a relatively small model size, low computational complexity, and short inference time. Note that introducing quadratic neurons in convolutional layers only increases the number of parameters moderately, because the number of parameters in fully-connected layers takes up a high percentage of the WDCNN. 

\begin{table*}[b]
\vspace{-0.2cm}
\caption{The average accuracy of all compared methods on noisy data. The bold number denotes the best performance.}
\scalebox{0.9}{
\begin{tabular}{|c|l|c|c|c|c|c|c|c|c|}
\hline
    & SNR & -6  & -4   & -2  & 0  & 2    & 4 & 6 & Average                                                                \\ \hline
\multirow{6}{*}{\makecell{CWRU\\-0HP}}  & DCA-BiGRU    & 0.6308\scriptsize $\pm$0.0065                                                          & 0.7164\scriptsize $\pm$0.0070                                                          & 0.7801\scriptsize $\pm$0.0079                                                          & 0.8572\scriptsize $\pm$0.0045                                                          & 0.8987\scriptsize $\pm$0.0065                                                          & 0.9480\scriptsize $\pm$0.0035                                                          & 0.9709\scriptsize $\pm$0.0024                                                          & 0.8288\scriptsize $\pm$0.0899                                                          \\ \cline{2-10} 
                      & AResNet      & 0.6121\scriptsize $\pm$0.0060                                                          & 0.6925\scriptsize $\pm$0.0044                                                          & 0.7712\scriptsize $\pm$0.0065                                                          & 0.8429\scriptsize $\pm$0.0073                                                          & 0.9106\scriptsize $\pm$0.0065                                                          & 0.9557\scriptsize $\pm$0.0029                                                          & 0.9785\scriptsize $\pm$.00033                                                          & 0.8234\scriptsize $\pm$0.1277                                                          \\ \cline{2-10} 
                      & RNN-WDCNN    & 0.5804\scriptsize $\pm$0.0059                                                          & 0.7274\scriptsize $\pm$0.0085                                                          & 0.8876\scriptsize $\pm$0.0070                                                          & 0.9549\scriptsize $\pm$0.0057                                                          & \textbf{0.9893\scriptsize $\pm$0.0018}               & 0.9938\scriptsize $\pm$0.0010                                                           & 0.9926\scriptsize $\pm$0.0015                                                          & 0.8751\scriptsize $\pm$0.1493                                                          \\ \cline{2-10} 
                      & MA1DCNN      & 0.6954\scriptsize $\pm$0.0069                                                          & 0.7435\scriptsize $\pm$0.0071                                                          & 0.8674\scriptsize $\pm$0.0052                                                          & 0.9319\scriptsize $\pm$0.0048                                                          & 0.9597\scriptsize $\pm$0.0025                                                          & 0.9749\scriptsize $\pm$0.0028                                                          & 0.9816\scriptsize $\pm$0.0021                                                          & 0.8792\scriptsize $\pm$0.1077                                                          \\ \cline{2-10} 
                      & WDCNN        & 0.6652\scriptsize $\pm$0.0056                           & 0.7736\scriptsize $\pm$0.0094                           & 0.8712\scriptsize $\pm$0.0022                           & 0.9420\scriptsize $\pm$0.0032                           & 0.9836\scriptsize $\pm$0.0056                           & 0.9952\scriptsize $\pm$0.0084                           & 0.9992\scriptsize $\pm$0.0003                           & 0.8900\scriptsize $\pm$0.1189                           \\ \cline{2-10} 
                      & QCNN         & \textbf{0.7112\scriptsize $\pm$0.0087} & \textbf{0.7992\scriptsize $\pm$0.0084} & \textbf{0.8928\scriptsize $\pm$0.0051} & \textbf{0.9576\scriptsize $\pm$0.0064} & 0.9852\scriptsize $\pm$0.0052 & \textbf{0.9964\scriptsize $\pm$0.0016} & \textbf{0.9996\scriptsize $\pm$0.0005} & \textbf{0.9060\scriptsize $\pm$0.1038} \\ \hline \hline
\multirow{6}{*}{\makecell{CWRU\\-1HP}}  & DCA-BiGRU    & 0.6374\scriptsize $\pm$0.0065                                                          & 0.7176\scriptsize $\pm$0.0057                                                          & 0.7901\scriptsize $\pm$0.0064                                                          & 0.8880\scriptsize $\pm$0.0050                                                          & 0.9418\scriptsize $\pm$0.0032                                                          & 0.9751\scriptsize $\pm$0.0026                                                          & 0.9784\scriptsize $\pm$0.0022                                                          & 0.8818\scriptsize $\pm$0.0975                                                          \\ \cline{2-10} 
                      & AResNet      & 0.6133\scriptsize $\pm$0.0054                                                          & 0.7046\scriptsize $\pm$0.0070                                                          & 0.7887\scriptsize $\pm$0.0043                                                          & 0.8732\scriptsize $\pm$0.0046                                                          & 0.9268\scriptsize $\pm$0.0039                                                          & 0.9684\scriptsize $\pm$0.0032                                                          & 0.9894\scriptsize $\pm$0.0016                                                          & 0.8378\scriptsize $\pm$0.1308                                                          \\ \cline{2-10} 
                      & RNN-WDCNN    & 0.5854\scriptsize $\pm$0.0079                                                          & 0.7422\scriptsize $\pm$0.0062                                                          & 0.8768\scriptsize $\pm$0.0050                                                          & \textbf{0.9658\scriptsize $\pm$0.0030}                                                           & 0.9776\scriptsize $\pm$0.0022                                                          & 0.9827\scriptsize $\pm$0.0019                                                          & 0.9846\scriptsize $\pm$0.0020                                                          & 0.8714\scriptsize $\pm$0.1434                                                          \\ \cline{2-10} 
                      & MA1DCNN      & \textbf{0.7308\scriptsize $\pm$0.0086}                                                           & 0.8286\scriptsize $\pm$0.0061                                                          & 0.9047\scriptsize $\pm$0.0047                                                          & 0.9622\scriptsize $\pm$0.0027                                                          & 0.9773\scriptsize $\pm$0.0027                                                          & 0.9853\scriptsize $\pm$0.0019                                                          & 0.9933\scriptsize $\pm$0.0016                                                          & 0.9117\scriptsize $\pm$0.0916                                                          \\ \cline{2-10} 
                      & WDCNN        & 0.6968\scriptsize $\pm$0.0051                           & 0.7268\scriptsize $\pm$0.0091                           & 0.8860\scriptsize $\pm$0.0017                           & 0.9228\scriptsize $\pm$0.0056                           & 0.9812\scriptsize $\pm$0.0036                           & 0.9792\scriptsize $\pm$0.0046                           & 0.9912\scriptsize $\pm$0.0007                           & 0.8834\scriptsize $\pm$0.1142                           \\ \cline{2-10} 
                      & QCNN         & 0.7080\scriptsize $\pm$0.0084 & \textbf{0.8144\scriptsize $\pm$0.0040} & \textbf{0.8992\scriptsize $\pm$0.0035} & 0.9624\scriptsize$\pm$0.0062& \textbf{0.9872\scriptsize $\pm$0.0078} & \textbf{0.9940\scriptsize $\pm$0.0079} & \textbf{0.9964\scriptsize $\pm$0.0071} & \textbf{0.9323\scriptsize $\pm$0.1023} \\ \hline \hline
\multirow{6}{*}{\makecell{CWRU\\-2HP}}  & DCA-BiGRU    & 0.6622\scriptsize $\pm$0.0055                                                          & 0.7477\scriptsize $\pm$0.0059                                                          & 0.8214\scriptsize $\pm$0.0051                                                          & 0.9148\scriptsize $\pm$0.0042                                                          & 0.9536\scriptsize $\pm$0.0030                                                          & 0.9719\scriptsize $\pm$0.0024                                                          & 0.9914\scriptsize $\pm$0.0019                                                          & 0.8661\scriptsize $\pm$0.0876                                                          \\ \cline{2-10} 
                      & AResNet      & 0.6098\scriptsize $\pm$0.0062                                                          & 0.7208\scriptsize $\pm$0.0050                                                          & 0.8191\scriptsize $\pm$0.0059                                                          & 0.8987\scriptsize $\pm$0.0041                                                          & 0.9550\scriptsize $\pm$0.0046                                                          & 0.9851\scriptsize $\pm$0.0020                                                          & 0.9961\scriptsize $\pm$0.0010                                                          & 0.8549\scriptsize $\pm$0.1353                                                          \\ \cline{2-10} 
                      & RNN-WDCNN    & 0.6790\scriptsize $\pm$0.0079                                                          & 0.8142\scriptsize $\pm$0.0051                                                          & 0.9322\scriptsize $\pm$0.0052                                                          & \textbf{0.9816\scriptsize $\pm$0.0020}                                                           & \textbf{0.9972\scriptsize $\pm$0.0009}                                                           & 0.9982\scriptsize $\pm$0.0006                                                          & 0.9999\scriptsize $\pm$0.0002                                                          & 0.9146\scriptsize $\pm$0.1143                                                          \\ \cline{2-10} 
                      & MA1DCNN      & 0.7587\scriptsize $\pm$0.0065                                                        & \textbf{0.8552\scriptsize $\pm$0.0083}                                                          & \textbf{0.9362\scriptsize $\pm$0.0052}                                        & 0.9747\scriptsize $\pm$0.0038                                                          & 0.9906\scriptsize $\pm$0.0017                                                          & 0.9970\scriptsize $\pm$0.0010                                                         & 0.9991\scriptsize $\pm$0.0005                                                          & 0.9302\scriptsize $\pm$0.0844                                                          \\ \cline{2-10} 
                      & WDCNN        & 0.7192\scriptsize $\pm$0.0033                           & 0.8372\scriptsize $\pm$0.0029                           & 0.9076\scriptsize $\pm$0.0052                           & 0.9780\scriptsize $\pm$0.0030 & 0.9908\scriptsize $\pm$0.0087                           & 0.9984\scriptsize $\pm$0.0044                           & 0.9996\scriptsize $\pm$0.0020                        & 0.9187\scriptsize $\pm$0.0987                           \\ \cline{2-10} 
                      & QCNN         & \textbf{0.7588\scriptsize $\pm$0.0095} & 0.8516\scriptsize $\pm$0.0066 & 0.9352\scriptsize $\pm$0.0060 & 0.9796\scriptsize $\pm$0.0063                           & 0.9964\scriptsize $\pm$0.0018 & \textbf{1.0000\scriptsize $\pm$0.0000} & \textbf{1.0000\scriptsize $\pm$0.0000}                           & \textbf{0.9323\scriptsize $\pm$0.0850} \\ \hline \hline
\multirow{6}{*}{\makecell{CWRU\\-3HP}}  & DCA-BiGRU    & 0.6653\scriptsize $\pm$0.0039                                                          & 0.7388\scriptsize $\pm$0.0071                                                          & 0.8328\scriptsize $\pm$0.0085                                                          & 0.8967\scriptsize $\pm$0.0069                                                          & 0.9528\scriptsize $\pm$0.0037                                                          & 0.9806\scriptsize $\pm$0.0022                                                          & 0.9925\scriptsize $\pm$0.0020                                                          & 0.8656\scriptsize $\pm$0.0898                                                          \\ \cline{2-10} 
                      & AResNet      & 0.6311\scriptsize $\pm$0.0069                                                          & 0.7244\scriptsize $\pm$0.0088                                                          & 0.8124\scriptsize $\pm$0..0081                                                         & 0.8962\scriptsize $\pm$0.0043                                                          & 0.9634\scriptsize $\pm$0.0025                                                          & 0.9917\scriptsize $\pm$0.0017                                                          & 0.9989\scriptsize $\pm$0.0006                                                          & 0.8597\scriptsize $\pm$0.1312                                                          \\ \cline{2-10} 
                      & RNN-WDCNN    & 0.6325\scriptsize $\pm$0.0085                                                          & 0.8094\scriptsize $\pm$0.0073                                                          & \textbf{0.9419\scriptsize $\pm$0.0039}                                                         & \textbf{0.9864\scriptsize$\pm$0.0021}                                                           & \textbf{0.9988\scriptsize$\pm$0.0006}                                                           & 0.9999\scriptsize $\pm$0.0001                                                          &  \textbf{1.0000\scriptsize $\pm$0.0000}                                                           & 0.9099\scriptsize $\pm$0.1299                                                          \\ \cline{2-10} 
                      & MA1DCNN      & \textbf{0.7463\scriptsize $\pm$0.0083}                                                           & 0.8391\scriptsize $\pm$0.0046                                                          & 0.9292\scriptsize $\pm$0.0054                                                          & 0.9747\scriptsize $\pm$0.0030                                                          & 0.9890\scriptsize $\pm$0.0029                                                          & 0.9945\scriptsize $\pm$0.0014                                                          & 0.9995\scriptsize $\pm$0.0006                                                          & 0.9246\scriptsize $\pm$0.0897                                                          \\ \cline{2-10} 
                      & WDCNN        & 0.7240\scriptsize $\pm$0.0085                           & 0.8264\scriptsize $\pm$0.0048                           & 0.9068\scriptsize $\pm$0.0037                           & 0.9660\scriptsize $\pm$0.0038                           & 0.9968\scriptsize $\pm$0.0024                           & 0.9988\scriptsize $\pm$0.0016                           & \textbf{1.0000\scriptsize $\pm$0.0000}                            & 0.9170\scriptsize $\pm$0.0985                           \\ \cline{2-10} 
                      & QCNN         & 0.7396\scriptsize $\pm$0.0093 & \textbf{0.8444\scriptsize $\pm$0.0046} & 0.9268\scriptsize $\pm$0.0043 & 0.9756\scriptsize $\pm$0.0014 & 0.9952\scriptsize $\pm$0.0027 & \textbf{1.0000\scriptsize $\pm$0.0000} & \textbf{1.0000\scriptsize $\pm$0.0000}                         & \textbf{0.9259\scriptsize $\pm$0.0923} \\ \hline \hline
\multirow{6}{*}{HIT} & DCA-BiGRU    & 0.3305\scriptsize $\pm$0.0087                                                          & 0.4742\scriptsize $\pm$0.0094                                                          & 0.6382\scriptsize $\pm$0.0105                                                          & 0.7442\scriptsize $\pm$0.0086                                                          & 0.8479\scriptsize $\pm$0.0045                                                          & 0.8896\scriptsize $\pm$0.0050                                                          & 0.9242\scriptsize $\pm$0.0030                                                          & 0.6927\scriptsize $\pm$0.2073                                                          \\ \cline{2-10} 
                      & AResNet      & 0.3705\scriptsize $\pm$0.0073                                                          & 0.4903\scriptsize $\pm$0.0106                                                          & 0.6642\scriptsize $\pm$0.0076                                                          & 0.8012\scriptsize $\pm$0.0075                                                          & 0.8807\scriptsize $\pm$0.0048                                                          & 0.9300\scriptsize $\pm$0.0051                                                          & 0.9512\scriptsize $\pm$0.0048                                                          & 0.7269\scriptsize $\pm$0.2096                                                          \\ \cline{2-10} 
                      & RNN-WDCNN    & 0.3598\scriptsize $\pm$0.0264                                                          & 0.4786\scriptsize $\pm$0.0521                                                          & 0.5922\scriptsize $\pm$0.0801                                                          & 0.6660\scriptsize $\pm$0.1026                                                          & 0.7224\scriptsize $\pm$0.1290                                                          & 0.7628\scriptsize $\pm$0.1378                                                          & 0.8013\scriptsize $\pm$0.1434                                                          & 0.6268\scriptsize $\pm$0.1483                                                          \\ \cline{2-10} 
                      & MA1DCNN      & 0.3467\scriptsize $\pm$0.0235                                                          & 0.4872\scriptsize $\pm$0.0517                                                          & 0.5914\scriptsize $\pm$0.0842                                                          & 0.7717\scriptsize $\pm$0.1152                                                          & 0.8305\scriptsize $\pm$0.1122                                                          & 0.8718\scriptsize $\pm$0.1030                                                          & 0.8644\scriptsize $\pm$0.1122                                                          & 0.6805\scriptsize $\pm$0.1920                                                          \\ \cline{2-10} 
                      & WDCNN        & 0.3940\scriptsize $\pm$0.0044                            & 0.5148\scriptsize $\pm$0.0026                           & 0.6308\scriptsize $\pm$0.0079                           & 0.7308\scriptsize $\pm$0.0024                            & 0.8444\scriptsize $\pm$0.0052                           & 0.8876\scriptsize $\pm$0.0092                           & 0.9288\scriptsize $\pm$0.0084                           & 0.7045\scriptsize $\pm$0.1859                           \\ \cline{2-10} 
                      & QCNN         & \textbf{0.4244\scriptsize $\pm$0.0013} & \textbf{0.5348\scriptsize $\pm$0.0087} & \textbf{0.6892\scriptsize $\pm$0.0020} & \textbf{0.8356\scriptsize $\pm$0.0033} & \textbf{0.8728\scriptsize $\pm$0.0082} & \textbf{0.9192\scriptsize $\pm$0.0007} & \textbf{0.9512\scriptsize $\pm$0.0061} & \textbf{0.7467\scriptsize $\pm$0.1880} \\ \hline
\end{tabular}}
\label{tab:noise}
\end{table*}

We set seven noise levels. The results are shown in Table \ref{tab:noise}. All results are the average and std of ten runs. We draw the following highlights from Table \ref{tab:noise}. First, for all datasets and SNR conditions, the QCNN outperforms its competitors in terms of average accuracy. In particular, on our and CWRU-1HP datasets, the average accuracy of QCNN is about 2\% higher than the second-best method and 4\% higher than the third. Although in some cases where the QCNN is not the best, it ranks second and admits a small gap to the first place. Therefore, we conclude that the QCNN is a versatile model that consistently delivers competitive performance at various noise levels. 

\begin{figure}[htbp]
    \centering
    \subfloat[QCNN]{\includegraphics[width=0.5\linewidth]{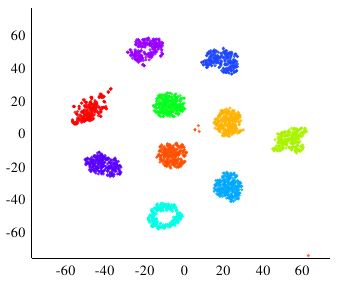}}
    \subfloat[WDCNN]{\includegraphics[width=0.5\linewidth]{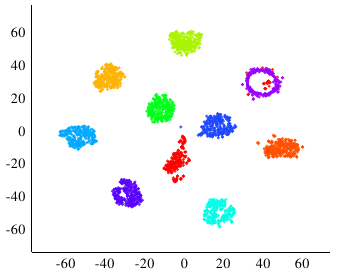}}
    \caption{The t-SNE results of features produced by the last convolutional layers of QCNN and WDCNN on CWRU-2HP dataset.}
    \label{fig:tsne}
\end{figure}

Second, in most cases, the QCNN outcompetes WDCNN. Particularly, the QCNN leads by a large margin on severely noisy signals, \textit{e.g.}, in SNR = -6dB CWRU-1HP dataset, the average accuracy of the QCNN is 4.4\% higher than the WDCNN. Since the QCNN and WDCNN share the same structure, it confirms that a quadratic neuron is a simple and effective way of augmenting a network's performance. To further compare what the QCNN and WDCNN learn from data, we use t-SNE \cite{van2008visualizing} to visualize output features of the last convolutional layers of the QCNN and WDCNN. As shown in Figure \ref{fig:tsne}, different colors denote different categories of bearings. Although both models exhibit promising feature extraction capabilities, evidenced by discernible clusters of 10 categories of data, the QCNN shows better feature discriminatory ability. For example, the purple cluster in the result of the WDCNN is contaminated by red points, but the purple and red clusters are well separated in the result of the QCNN. 

\begin{figure}[htbp]
    \centering
    \includegraphics[width=\linewidth]{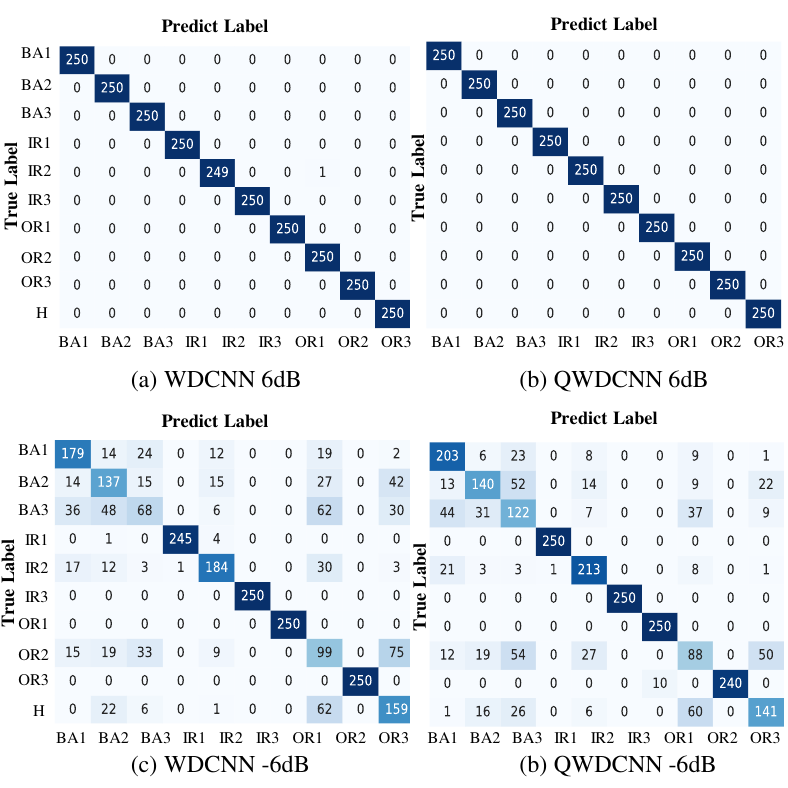}
    \caption{Confusion matrix of QCNN and WDCNN on CWRU-2HP. BA, IR, OR, and H denote ball faulty, inner race faulty, outer race faulty, and healthy, respectively.}
    \label{fig:confusion}
\end{figure}

At last, we calculate confusion matrices of WDCNN and QWDCNN to analyze the classification performance of each fault mode in detail. As shown in Figure \ref{fig:confusion} (a) and (b), both QCNN and WDCNN shows excellent performance under weak noisy condition, WDCNN still misjudges one sample, while QCNN is completely correct. Moreover, the classification results of WDCNN and QCNN on strong noise are exhibited in Figure \ref{fig:confusion} (c)-(d). Most misclassified samples come from ball faults, because the ball fault characteristic frequency is modulated by the cage speed, and both are also accompanied by random spin.\cite{randall2011rolling}. As a result, random interference introduced by strong noise makes it difficult to identify different levels of ball faults. Compared to WDCNN and QCNN, QCNN correctly identifies a larger number of ball faults than WDCNN. In particular, if we consider all ball faults holistically, the diagnostic accuracy of QCNN is higher ($634/750=84.53\%$) than that of WDCNN( $535/750=71.33\%$). Finally, We are more interested in the number of faulty samples judged to be healthy. That is, the last column of the confusion matrix. The QCNN misjudges 83 samples, while the WDCNN is 149 samples. All results suggest that QCNN has better performance in noisy conditions.

\begin{figure*}[t]
    \includegraphics{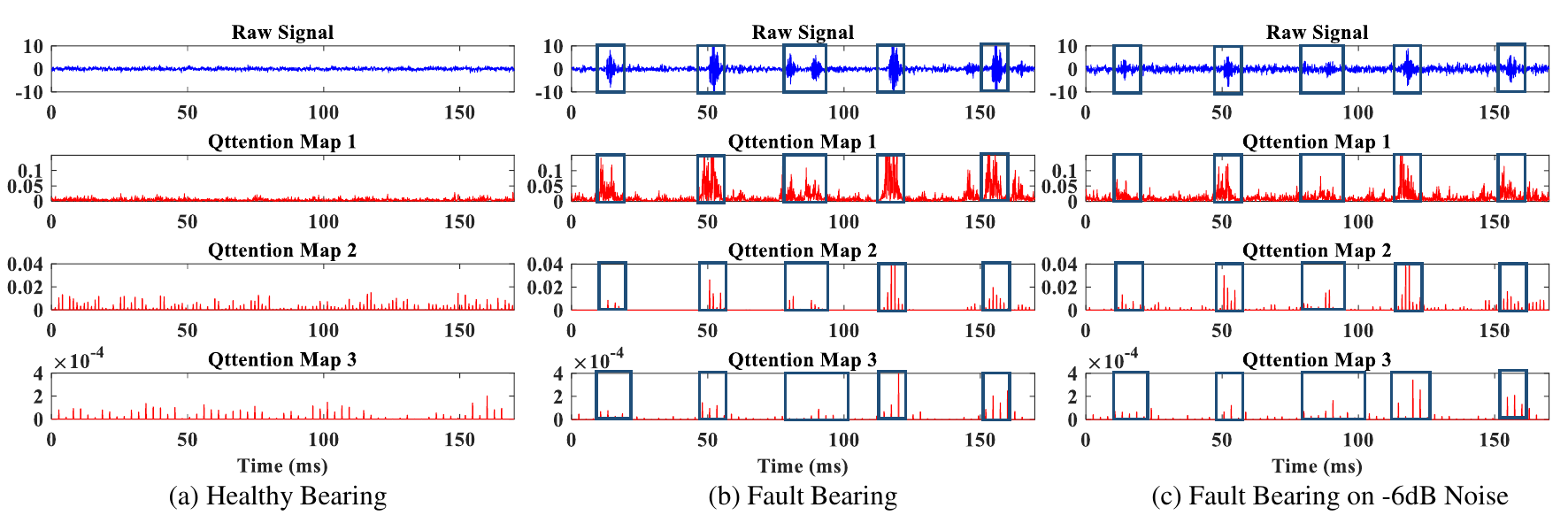}
    \caption{Three convolutional layers' qttention maps on CWRU 1HP. From shallow to deep, the network gradually intensifies its focus on the strongest vibration features.}
    \label{fig:att1}
    \vspace{-0.5cm}
\end{figure*}



\subsection{Interpretability of Qttention}


Interpretability of deep learning can be divided into model explainability and data explainability. Visualisation of the intrinsic weights of a deep learning model is considered an understanding of the model learning process, offering an intuitive evaluation of the model's performance \cite{sun2022interpretable}. In the bearing faults diagnosis, We can obtain the interpretability of the attention module through some visualisation techniques, \textit{i.e,} CAM \cite{zhou2016learning}, Grad-CAM \cite{selvaraju2017grad}, Grad-CAM++ \cite{chattopadhay2018grad, zhang2022fault}. Previous works have shown that the attention mechanism has the property of orienting the model to focus on key features of the data \cite{wang2019understanding, chen2022fault, yang2020interpreting, li2019understanding}. As we deduced before, qttention is a quadratic neuron-induced attention mechanism of showing what a convolutional layer focuses on from input signals. The values of the qttention represent the importance levels of the input $\boldsymbol{x}$ so that a quadratic network admits a self-explanatory feature extraction capability. This is a unique advantage of using quadratic neurons. In contrast, the conventional neuron doesn't enjoy such a kind of explainability because the linear coefficients are independent of the input, and cannot serve as the attention map.

\begin{figure}[htbp]
    \centering
    \includegraphics[width=\linewidth]{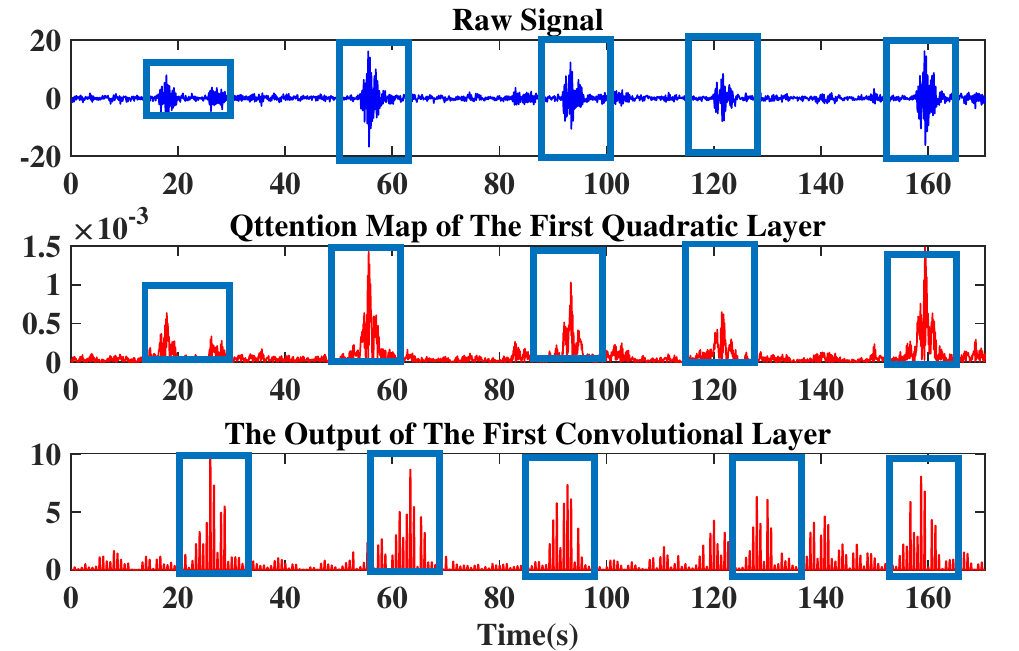}
    \caption{Comparison of qttention map and the output of traditional convolutional layer.}
    \label{fig:qvc}
\end{figure}

\underline{The comparisons between qttention and convolution.} Here, we give a comparison of the qttention map and the analogue operation of a conventional convolutional layer. We compute the output using Eq.(\ref{eq:conventional_attention})

\begin{equation}
    Out(\boldsymbol{x}) = |\mathrm{Grad}(\boldsymbol{x}^{\top}\boldsymbol{w})|.
\label{eq:conventional_attention}
\end{equation}
The results are shown in Figure \ref{fig:qvc}. The qttention map of the first layer more accurately conforms to the raw signal. This means that the quadratic network notices where the vibration intensity of the signal is higher. In contrast, the conventional network has larger weight values around high amplitude signals, but is not aligned with the raw signal and contains interference noise. Therefore, it lacks interpretability of the features extracted from the network.

Furthermore, we conduct a classification experiment to verify whether a qttention map indeed captures important faults feature. We use a qttention map and the output of the first convolutional layer as the input of an SVM classifier. We feed the original signals directly into an SVM classifier as a baseline. As Table \ref{tab:qttention_classify} shows, the highlight is that QCNN+SVM outperforms others by a large margin, suggesting that the qttention map extracts bearing fault features to enhance the classification of a simple classifier. Compared to a baseline SVM, a QCNN+SVM shows a significant performance improvement. In particular, on the HIT dataset, QCNN+SVM is 14.64\% higher than SVM. What's more, note that WDCNN+SVM has a surprisingly low classification accuracy. This phenomenon indicates that conventional neural networks may not identify as important bearing fault features as qttention does. This also agrees with the visualisation in Figure \ref{fig:qvc}. Hence, we argue that a quadratic network is an inherently more interpretable model than a conventional network.
\begin{figure*}[h]
    \centering
    \includegraphics{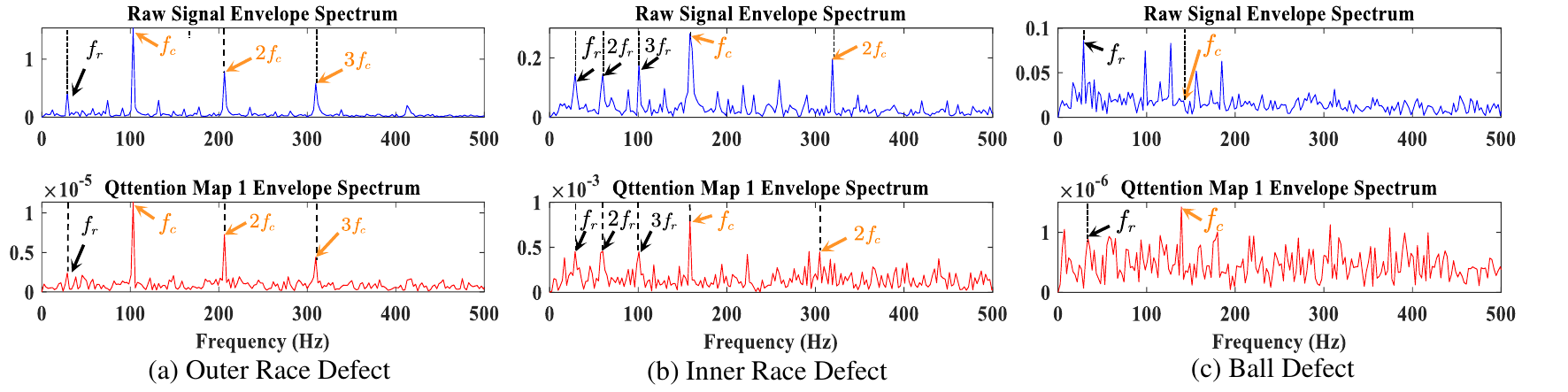}
    \caption{The envelope spectra (frequency spectrum of the Hilbert-Huang transform) of the raw signals and the qttention maps. $f_r$ denotes shaft characteristic frequency, $f_c$ denotes characteristic frequency of different defects.}
    \label{fig:att2}
    \vspace{-0.6cm}
\end{figure*}

\begin{table}[htbp]
\centering
  \caption{The average accuracy of all methods on the original datasets. The bold-faced numbers denote the best performance. WDCNN+SVM denotes using the first layer output of a WDCNN to train an SVM classifier, and QCNN+SVM denotes using the first layer qttention map.}
\begin{tabular}{|l|c|c|c|}
\hline
Datasets & SVM    & WDCNN+SVM & QCNN+SVM \\ \hline
CWRU-0HP & 0.8372 & 0.0980    & \textbf{0.9020}    \\ \hline
CWRU-1HP & 0.7960 & 0.0960    & \textbf{0.8956}    \\ \hline
CWRU-2HP & 0.7828 & 0.1012    & \textbf{0.9116}    \\ \hline
CWRU-3HP & 0.8708 & 0.1024    & \textbf{0.9176}    \\ \hline
HIT      & 0.6636 & 0.0996    & \textbf{0.8100}    \\ \hline
\end{tabular}
\label{tab:qttention_classify}
\end{table}

\underline{Visualization in time domain.} Qttention informs what local information a quadratic convolutional layer highlights and the flow of important information across layers. Here, we compute qttention maps from the earlier three quadratic convolutional layers for healthy, faulty, and noisy faulty bearing signals. Because each convolutional layer is followed by a max-pooling layer, the lengths of signals, as well as the corresponding qttention maps, become short. So at the second and third layers, we need to interpolate the attention map to the same dimension of the input. Figure \ref{fig:att1} showcases the qttention maps of three signals, where a higher value of qttention means higher importance. First, comparing healthy bearing and faulty bearing, qttention values of the healthy bearing are lower, suggesting that the QCNN pays more attention to the faulty signal. Second, for faulty bearing, the qttention shares a similar trend with the raw signal. All high-magnitude sequences, where the faulty area might be contacted to generate a severe vibration, are captured. As the layer goes deep, the attention to key faulty sequences retains. At last, in regards to the noisy faulty bearing signal, the qttention map can still attend the faulty positions even when the noise almost dominates the raw signal. Moreover, the qttention map of the third convolutional layer keeps similar features to that of the noiseless faulty bearing. This phenomenon accounts for why the QCNN can maintain good performance for noisy signals; the QCNN can consistently acquire useful features even if the signal is noisy.

\underline{Visualization in envelope spectrum.} The bearing vibration signal is mixed with multiple frequency components. The sign of a bearing defect is a significant impulse in the frequency domain (the corresponding frequency is called the characteristic frequency). While in the low-frequency domain, there exist the shaft frequency and its harmonics. Although we show that the qttention map can exact fault features to enhance the downstream classifier, we are curious to the questions: i) Does the QCNN learn features consistent with the physical interpretation? ii) What is the inherent mechanism of the QCNN for learning fault features? To resolve these questions, we validate the envelope spectrum of the qttention map of the first convolutional layer by considering the physics principle of bearing defect vibration signals.

The characteristic frequency of bearing defects is caused by balls passing through defect points. Given that $n, d$, and $D$ are the number of balls, the ball diameter, and the bearing pitch diameter, $f_r$ is shaft speed, and $\phi$ is the angle of the load from the radial plane, the characteristic frequencies of bearing defects obey the following equations:
\begin{equation}
\begin{cases}
&f_{\rm BPFO}=\frac{nf_r}{2}\left( 1-\frac{d}{D}\cos \phi \right)
\\
&f_{\rm BPFI}=\frac{nf_r}{2}\left( 1+\frac{d}{D}\cos \phi \right)
\\
&f_{\rm FTF}=\frac{f_r}{2}\left( 1-\frac{d}{D}\cos \phi \right)
\\
&f_{\rm BSF}=\frac{Df_r}{2d}\left( 1-\left( \frac{d}{D}\cos \phi \right) ^2 \right),
\end{cases}
\end{equation}
where $f_{\rm BPFO}, f_{\rm BPFI}$ is the ball pass frequency of the outer race and inner race, $f_{\rm FTF}$ is the fundamental train frequency (cage speed), and $f_{\rm BSF}$ is the ball spin frequency \cite{randall2011rolling, smith2015rolling}. 

Hilbert-Huang transform (HHT) is a powerful technique to analyze non-stationary signals \cite{randall2011rolling}. By extracting the envelope of the signal through HHT and calculating its frequency spectrum, the characteristic frequencies of the non-stationary signal can be clearly observed \cite{rai2007bearing}. In \cite{wang2018bearing} the characteristic frequencies of faults in the CWRU dataset are computed, here we adopt those computations. Figure \ref{fig:att2} shows the envelope spectra of the raw signals and qttention maps. We use $f_c$ to uniformly represent $f_{\rm BPFO}, f_{\rm BPFI}, f_{\rm BSF}$ for simplicity. First, both the envelope spectra of the outer race defect and inner race defect signals exhibit significant shaft frequency and characteristic frequencies, but the characteristic frequency of ball defects is hard to find because the ball is accompanied by spinning and rolling during motion. Second, by comparing the envelope spectra of the raw signal and the qttention map, we find that the QCNN favors bearing fault characteristics over the shaft. For the outer race defect, $f_c, 2f_c,$ and $3f_c$ are maintained while the shaft frequency is suppressed. For the inner race, $f_c$ and $2f_c$ are maintained, while $3f_c$ are extracted from the raw signal. For ball defects, the $f_c$ element is highly attended, whereas the shaft frequency is moderately attended. We argue that the QCNN indeed learns faulty features via working as an adaptive band-pass filter for different fault modes. At last, regarding the ball defect signal, when the characteristic frequency of the fault is interfered with by signals in similar frequency bands, the QCNN can still accurately identify the characteristic frequency $f_c$, which confirms that the QCNN has a powerful feature extraction capability.

\section{Conclusions}
\label{sec:Conclusions}

In this article, we have proposed a convolutional network made of recently-developed quadratic neurons for the end-to-end time-domain bearing fault diagnosis. Moreover, not only the proposed quadratic convolutional network can have competitive results on two bearing faults datasets, but also we have independently derived that a quadratic neuron inherently contains an attention mechanism, referred to as qttention. Furthermore, we have discussed differences between qttention and CNNs, and have conducted a classification test to verify that the bearing fault features attended by qttention are indeed important. We have utilized the qttention map to interpret the features learned by the proposed quadratic model in both time and frequency domains and elucidate the physics principle behind. In the future, more efforts are required to refine the qttention mechanism and find real-world applications.

\section*{Acknowledgments}
The experiments of this work is supported by Prof. Xiaoli Zhao, Mr. Hongyuan Zhang in MIIT Key Laboratory of Aerospace Bearing Technology and Equipment, Harbin Insti-
tute of Technology.



 
%
\bibliographystyle{IEEEtran}
\bibliography{bio.bib}





\vfill

\end{document}


\title{Supplementary Materials of "Quadratic Neuron-Induced Attention (Qttention) for Effective and Interpretable Bearing Fault Diagnosis"}

\author{Jing-Xiao~Liao$^{1}$, Zhi-Qi Sun$^{1}$, Hang-Cheng Dong, Jinwei Sun$^{1}$,  Shiping Zhang$^{1*}$, Feng-Lei Fan$^{2*}$
\thanks{*Shiping Zhang (spzhang@hit.edu.cn) and Feng-Lei Fan (hitfanfenglei@gmail.com) are co-corresponding authors.}
\thanks{$^{1}$Jing-Xiao Liao, Zhi-Qi Sun, Hang-Cheng Dong, Jinwei Sun, Shiping Zhang are with School of Instrumentation Science and Engineering, Harbin Institute of Technology, Harbin, China.}
\thanks{$^{2}$Feng-Lei Fan is with Weill Cornell Medicine, Cornell University, New York City, USA.}}



\maketitle

\section{Ablation study}
We conduct two experiments to study the property of a quadratic network. Moreover, the ablation study demonstrates that the proposed QCNN has outperformed other styles of quadratic neuron based models and a conventional neural network.

Compared to the baseline method WDCNN, our proposed method replaces conventional convolution operation with a quadratic convolution and keeps conventional neurons in fully-connected layers. Here, we conduct an ablation study to investigate the effects of all techniques. Classification experiments with no noise and an SNR = 0 dB are performed on our dataset. We modify the equation of a quadratic neuron and observe the effect of each on the classification results. Here, QCNN-ng denotes that a quadratic neuron does not have $\boldsymbol{x}^\top\boldsymbol{w}^{g}+b^{g}$ term. The equation is as follows
\begin{equation}
\nonumber
\sigma(f_{ng}(\boldsymbol{x})) = \sigma\Big((\boldsymbol{x}^\top\boldsymbol{w}^{r}+b^{r})+(\boldsymbol{x}\odot\boldsymbol{x})^\top\boldsymbol{w}^{b}+c\Big),
\end{equation}
QCNN-np denotes a quadratic neuron does not have power term. Such that
\begin{equation}
\nonumber
\sigma(f_{np}(\boldsymbol{x})) = \sigma\Big((\boldsymbol{x}^\top\boldsymbol{w}^{r}+b^{r})(\boldsymbol{x}^\top\boldsymbol{w}^{g}+b^{g})\Big).
\end{equation}
QCNN-aq denotes all neurons (convolutional layers and fully-connected layers) are quadratic neurons and QCNN-base is the structure we proposed.

\begin{figure}[htbp]
    \centering
    \subfloat[No Noise]{\includegraphics[width=\linewidth]{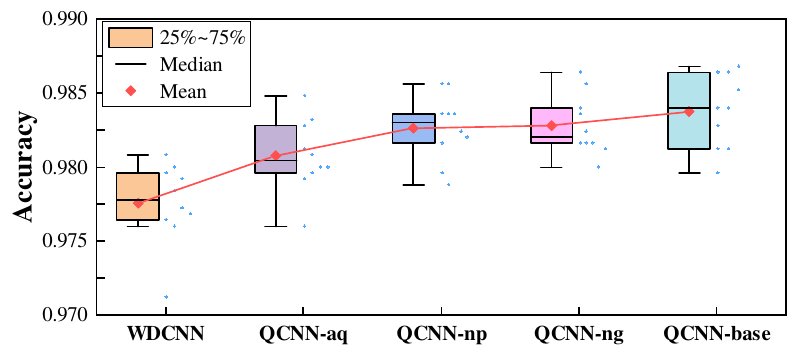}}\\
    \subfloat[0dB Noise]{\includegraphics[width=\linewidth]{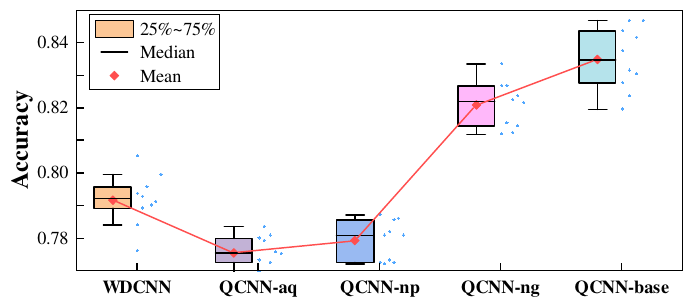}}
    \caption{The box-plot of compared structure on our dataset. A QCNN-aq denotes all quadratic neurons. A QCNN-np denotes no power term and a QCNN-ng denotes no $g$ term in a quadratic neuron. QCNN-base is the proposed method.}
    \label{fig:ablation}
\end{figure}

Table \ref{tab:ablation} and Figure \ref{fig:ablation} shows the results. First, the proposed method outperforms others, in particular, QCNN-base shows significant advantages in noise environments, and the accuracy of a QCNN-base is 4.32\% higher than that of a WDCNN. Second, a full quadratic network (QCNN-aq) does not show competitive results, with a noise condition, it is even worse than a WDCNN. This phenomenon indicates that quadratic neurons in fully-connected layers are not easy to fine-tuning. Moreover, in a noise environment, it tends to learn the noise features of the train set, imposing severe overfitting and leading to poor results in the test set. At last, compared to a QCNN-np and QCNN-ng, their performances are similar in no noise conditions but a QCNN-ng shows superior results in noise conditions. The lack of one first-order item does not cause significant performance loss, while the power term of a quadratic neuron provides an important role in the network.

\begin{table}[htbp]
\centering
\caption{The average accuracy of compared structure on our dataset.}
\scalebox{0.9}{
\begin{tabular}{|l||c|c|c|c|c|}
\hline
SNR & WDCNN  & QCNN-aq & QCNN-np & QCNN-ng & QCNN-base \\ \hline
N/A  & 0.9776 & 0.9808  & 0.9826  & 0.9828  & \textbf{0.9838} \\ \hline
0dB & 0.7917 & 0.7754  & 0.7792  & 0.8209  & \textbf{0.8349} \\ \hline
\end{tabular}}
\label{tab:ablation}
\end{table}

\vfill